%% file: main.tex
\documentclass{article}

\usepackage{microtype}
\usepackage{graphicx}
\usepackage{subfigure}
\usepackage{booktabs} 
\usepackage{listings}

\usepackage{hyperref}

\usepackage[accepted]{icml2020_arxiv}
\def\isarxiv{}

\usepackage{amsmath}
\usepackage{amsfonts}

\input{macros}

\icmltitlerunning{Differentiable Implicit Soft-Body Physics}

\begin{document}

\twocolumn[
\icmltitle{Differentiable Implicit Soft-Body Physics}



\icmlsetsymbol{equal}{*}

\begin{icmlauthorlist}
\icmlauthor{Junior Rojas}{utah}
\icmlauthor{Eftychios Sifakis}{wisconsin}
\icmlauthor{Ladislav Kavan}{utah}
\end{icmlauthorlist}

\icmlaffiliation{utah}{University of Utah, Salt Lake City, UT, United States}
\icmlaffiliation{wisconsin}{University of Wisconsin-Madison, Madison, WI, United States}

\icmlcorrespondingauthor{Junior Rojas}{jrojasdavalos@gmail.com}

\icmlkeywords{Machine Learning, ICML}

\vskip 0.3in
]



\printAffiliationsAndNotice{}  

\input{abstract}

\input{00-intro}
\input{01-related-work}
\input{02-quasistatic}
\input{03-differentiable-quasistatic}
\input{04-dynamics}
\input{05-conclusion}
\input{acknowledgements}





\bibliography{main}
\bibliographystyle{icml2020}





\end{document}

%% file: macros.tex
\let\vc=\mathbf
\newcommand{\R}{\mathbb{R}} 
\newcommand{\T}{\mathsf T} 
\newcommand{\dpartial}[2]{\frac{\partial {#1}}{\partial {#2}}}
\newcommand{\dd}[2]{\frac{d {#1}}{d {#2}}}
\newcommand{\argmin}[2]{\underset{#1}{\mathrm{argmin}}{\ #2}}

\newcommand{\refig}[1]{Figure~\ref{fig:#1}}
\newcommand{\refeq}[1]{Equation~\ref{eq:#1}}
\newcommand{\refsec}[1]{Section~\ref{sec:#1}}

%% file: abstract.tex
\begin{abstract}
We present a differentiable soft-body physics simulator that can be composed with neural networks as a differentiable layer.
In contrast to other differentiable physics approaches that use explicit forward models to define state transitions, we focus on implicit state transitions defined via function minimization.
Implicit state transitions appear in implicit numerical integration methods, which offer the benefits of large time steps and excellent numerical stability, but require a special treatment to achieve differentiability due to the absence of an explicit differentiable forward pass.
In contrast to other implicit differentiation approaches that require explicit formulas for the force function and the force Jacobian matrix, we present an energy-based approach that allows us to compute these derivatives automatically and in a matrix-free fashion via reverse-mode automatic differentiation.
This allows for more flexibility and productivity when defining physical models and is particularly important in the context of neural network training, which often relies on reverse-mode automatic differentiation (backpropagation).
We demonstrate the effectiveness of our differentiable simulator in policy optimization for locomotion tasks and show that it achieves better sample efficiency than model-free reinforcement learning.
\end{abstract}

%% file: 00-intro.tex
\section{Introduction}

Physics engines have been extensively used in robotics and computer graphics, but the increased adoption of neural networks to solve inverse problems in these domains has revealed the need for additional features in such engines \cite{degrave-differentiable, hu2019chainqueen, hu2019-taichi-lang, difftaichi}.
For example, using neural networks to solve inverse problems where physical constraints must be satisfied, such as motor control, often requires model-free reinforcement learning algorithms that approximate the derivatives of the simulator by sampling multiple trajectories using stochastic control policies.
This is due to the non-differentiable nature of most existing simulators, which have traditionally focused on efficiency and stability of forward computations only.
Differentiable simulators are useful when we want to incorporate physics into neural networks as a differentiable layer, which allows us to run backpropagation as commonly done in supervised learning.

\begin{figure}[h]
    \centering
    \includegraphics[width=\linewidth]{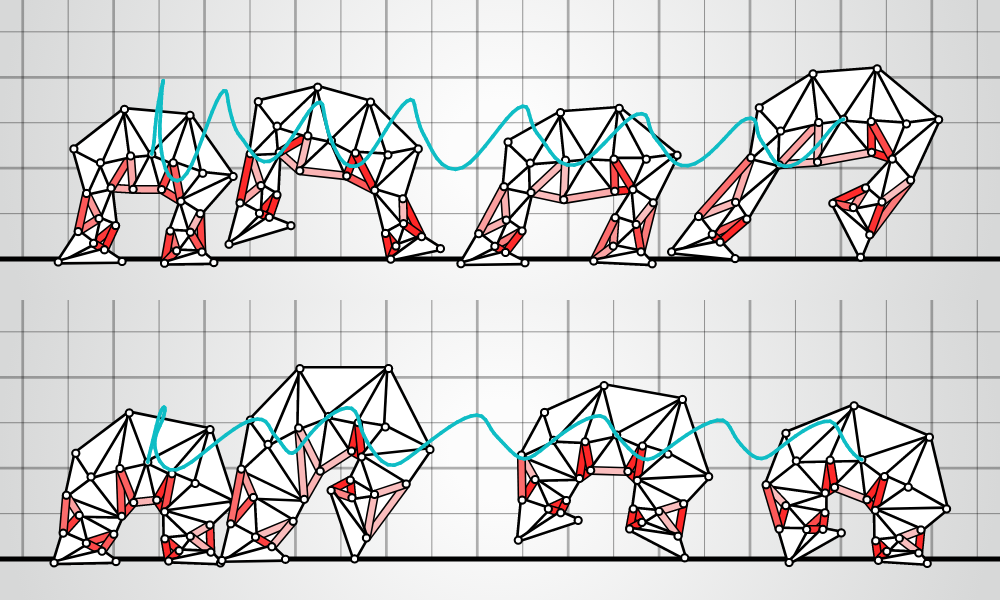}
    \caption{
        Locomotion control policies learned via backpropagation through time using our differentiable simulator.
        \ifdefined\isarxiv
        \else
            Videos of these motions can be found in the supplementary material.
        \fi
    }
    \label{fig:teaser}
\end{figure}

In this paper, we focus on soft-body physics and muscle-like actuation mechanisms, motivated by humans and other animals that control their bodies via storage and release of energy in their muscles to generate a wide range of motions, and we demonstrate that our differentiable simulator is effective in optimizing control policies for locomotion (\refig{teaser}) and can achieve better sample efficiency than model-free reinforcement learning.
In contrast to other differentiable soft-body physics approaches that use explicit forward models to define state transitions, we focus on implicit state transitions defined via function minimization.
Implicit state transitions appear in implicit numerical integration methods, which offer the benefits of large time steps and excellent numerical stability, but require a special treatment to achieve differentiability due to the absence of an explicit differentiable forward pass.

%% file: 01-related-work.tex
\section{Related work}

\subsection{Differentiable explicit soft-body physics}

Recent work \cite{hu2019chainqueen} showed that incorporating a differentiable physics engine, ChainQueen, into machine learning workflows can achieve better sample efficiency than model-free reinforcement learning in soft robotics tasks. Besides differentiability, the key characteristic of ChainQueen is its efficient GPU implementation of soft-body dynamics using the material point method and explicit numerical integration, which has been shown to have better runtime performance than an equivalent implementation in TensorFlow, whose numerical kernels are not as optimized for sparse operations.

This then led to the development of DiffTaichi \cite{difftaichi}, now part of the Taichi programming language \cite{hu2019-taichi-lang}, which introduced an optimizing compiler for differentiable sparse operations that commonly appear in physics-based simulation. The results presented in such line of work have focused on explicit numerical integration. An explicit differentiable forward pass is required in order to compute derivatives with DiffTaichi. Although standard deep learning packages can achieve differentiability of explicit operations too, they often lack efficient implementations of sparse operations, which is one of the main problems addressed by DiffTaichi.

However, explicit differentiation is not applicable to simulators that define implicit state transitions. For example, simulators that use implicit numerical integration offer the benefits of large time steps and better numerical stability \cite{gast-optim}, but do not define an explicit differentiable forward pass. The availability of efficient implementations of sparse operations has an important effect on the runtime performance of the simulator, but this aspect is orthogonal to the problem of achieving differentiability through implicit state transitions, which we tackle in this paper.

\subsection{Differentiable optimization}

The implicit state transitions that we focus on in this paper are primarily driven by 
optimization procedures. For example, quasistatic motions are driven by potential energy minimization. There has been recent work incorporating numerical optimization into neural networks as a differentiable layer \cite{amos2017optnet, cvxpylayers2019}. However, work in this area has focused on convex optimization, while the optimization problems that arise in soft-body physics are non-convex.

\subsection{Implicit layers}

Differentiable optimization can be considered part of a larger class of models sometimes referred to as implicit layers, which also include models such as neural ODEs \cite{neural-odes} and deep equilibrium models \cite{deqs}. These are layers whose output $y$ is not defined in terms of its input $x$ via some explicit rule $y = f(x)$, but rather implicitly via some specification $f(x, y) = 0$. Differentiating through implicit definitions requires special treatment and although there is previous work on implicit differentiation (often presented under the name of sensitivity analysis and adjoint method) for material parameter estimation and trajectory optimization in soft-body simulation \cite{sifakis2005automatic, bern2019trajectory, animated-plushies, bern-fabrication-modeling-control-plush-robots}, incorporating this implicit differentiation approach into neural networks for motor control and learning via backpropagation is less explored \cite{add-siggraph}. One common pattern in these previous approaches is that they use explicit formulas for the force function and the force Jacobian matrix. In this paper, however, we take an energy-based approach that allows us to compute these derivatives automatically using reverse-mode automatic differentiation, allowing for more flexibility and productivity when defining physical models.

\subsection{Energy-based models}

Energy-based modeling is a framework to define dependencies between variables via scalar-valued energy functions \cite{ebm-tutorial}. In these models, the forward pass is also defined implicitly, via energy minimization, and the term ``energy'' used in the context of energy-based modeling has a particularly close connection to the notion of potential energy in quasistatic motions that we consider in \refsec{quasistatic-motions}.

One of the main problems studied in energy-based modeling is how to learn energy functions, which can be modeled using neural networks.
However, since in this paper we consider problems in simulation where we already have access to the energy functions used to model the environment, we do not tackle this problem.
Instead, we focus on using our differentiable model to train control policies, but we emphasize that the energy functions that we use can be viewed as neural networks and our simulator is essentially a collection of energy functions implemented using operations commonly available in deep learning packages.
Additional quantities required for the forward and backward pass, such as forces and force Jacobians, can be derived from the energy function using automatic differentiation.
Our approach is especially relevant in the context of energy-based modeling, where the premise is that the primary definition should be the energy function and everything else is derived from such definition.

%% file: 02-quasistatic.tex
\section{Quasistatic motions}
\label{sec:quasistatic-motions}

We first consider quasistatic motions, which are primarily driven by potential energy minimization and ignore inertial effects, which we will consider in \refsec{dynamics}. While typical elastic objects are passive and their motion is determined by initial conditions, in this work we focus on active objects, which combine passive elasticity with an internal actuation mechanism such as muscles or robotic actuators.

\begin{figure}[h]
    \centering
    \includegraphics[width=\linewidth]{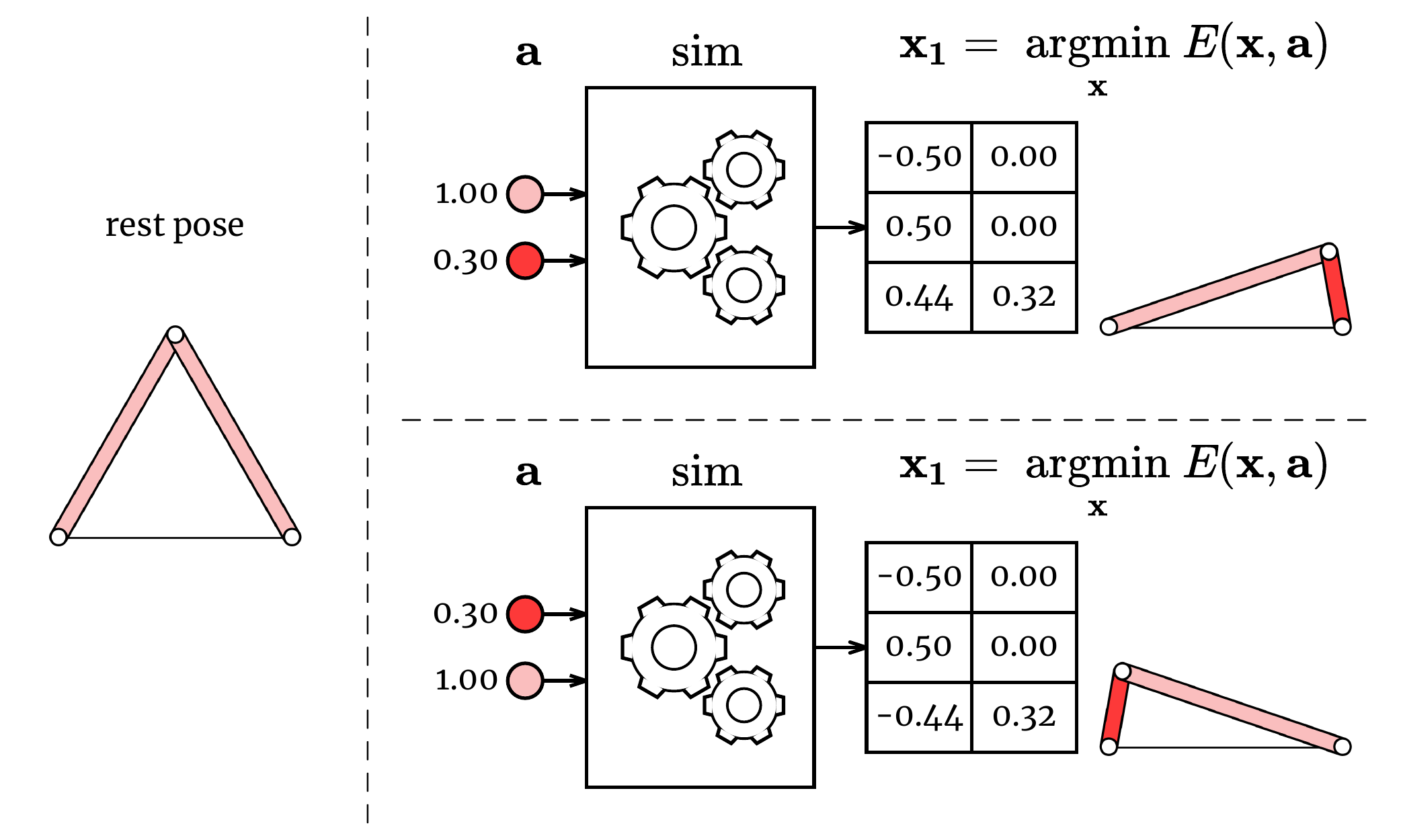}
    \caption{For didactic purposes, we consider a single 2D triangle with contractile fibers modeled as springs that can adjust their rest lengths via an action signal $\vc{a}$. In this model, an action value of 1 means that the fiber is trying to keep its original rest length, which is analogous to muscle relaxation, and values that are less than 1 mean that the fiber is trying to compress to some fraction of its original rest length, which is analogous to muscle contraction.}
    \label{fig:qs-action}
\end{figure}

The model we use for our experiments is based on deformable meshes simulated using the finite element method (FEM) and contractile fibers modeled as springs. As seen in \refig{qs-action}, the output of the simulator depends on the action signal $\vc{a}$. More specifically, it is computed by minimizing the potential energy $E$ of the system. For soft-body physics, the potential energy is conventionally defined as a function of vertex positions $\vc{x}$, but to support actuation mechanisms we consider energy functions $E$ that also take action signals $\vc{a}$ as input:

\begin{equation}
    \label{eq:x1-quasistatic-argmin}
    \vc{x_1}(\vc{a}) = \argmin{\vc{x}} E(\vc{x}, \vc{a})
\end{equation}

The energy function we use for each contractile fiber in the model shown in \refig{qs-action} is:

\begin{equation}
    \label{eq:active-spring-potential}
    E_{\mathrm{spring}}(\vc{x}, a) = \frac{k}{2} \left( \frac{l(\vc{x})}{a\ l_0} - 1 \right)^2
\end{equation}

where $l(\vc{x})$ is the current length of the fiber, $l_0$ is its original rest length and $k$ is its stiffness. However, we want to emphasize that the approach we propose to achieve differentiability is not tied to this particular model. The important characteristic of the model presented in this section is that it defines the output of the simulator via energy minimization (\refeq{x1-quasistatic-argmin}), but many possible energy functions exist for this purpose and our approach can achieve differentiability for arbitrary energy functions.

The energy function we use for our experiments also includes contributions from triangle deformations due to Neo-Hookean elasticity.
Although there are many variants of Neo-Hookean elasticity \cite{stable-neohookean}, the specific model we use is based on the energy density function $\Psi$ presented in \cite{femdefo}, which can be expressed in terms of the isotropic invariants $I_1$ and $J$ that can be computed from vertex positions $\vc{x}$:

\begin{equation}
    \label{eq:neohookean-psi}
    \Psi(I_1, J) = \frac{\mu}{2} (I_1 - 2) - \mu \log(J) + \frac{\lambda}{2} \log^2(J)
\end{equation}

The only modification we make to this model is that we use a quadratic approximation of the logarithmic function to avoid issues with non-positive values, which may appear when triangles are inverted \cite{irving-invertible}.
We refer to \cite{femdefo} for additional details on how to compute $I_1$ and $J$ from vertex positions $\vc{x}$ and additional background in FEM simulation.
However, we want to emphasize again that our goal in this paper is not to advocate any specific elasticity model or actuation mechanism.
We provide some details of the energy functions we use for our experiments for reference, but our goal is to provide a general framework to achieve differentiability in soft-body simulations where the forward pass requires a minimization procedure.
The specific function that the simulator minimizes and the minimization algorithm implemented in the simulator may vary depending on application requirements.

\begin{figure}[h]
    \centering
    \includegraphics[width=\linewidth]{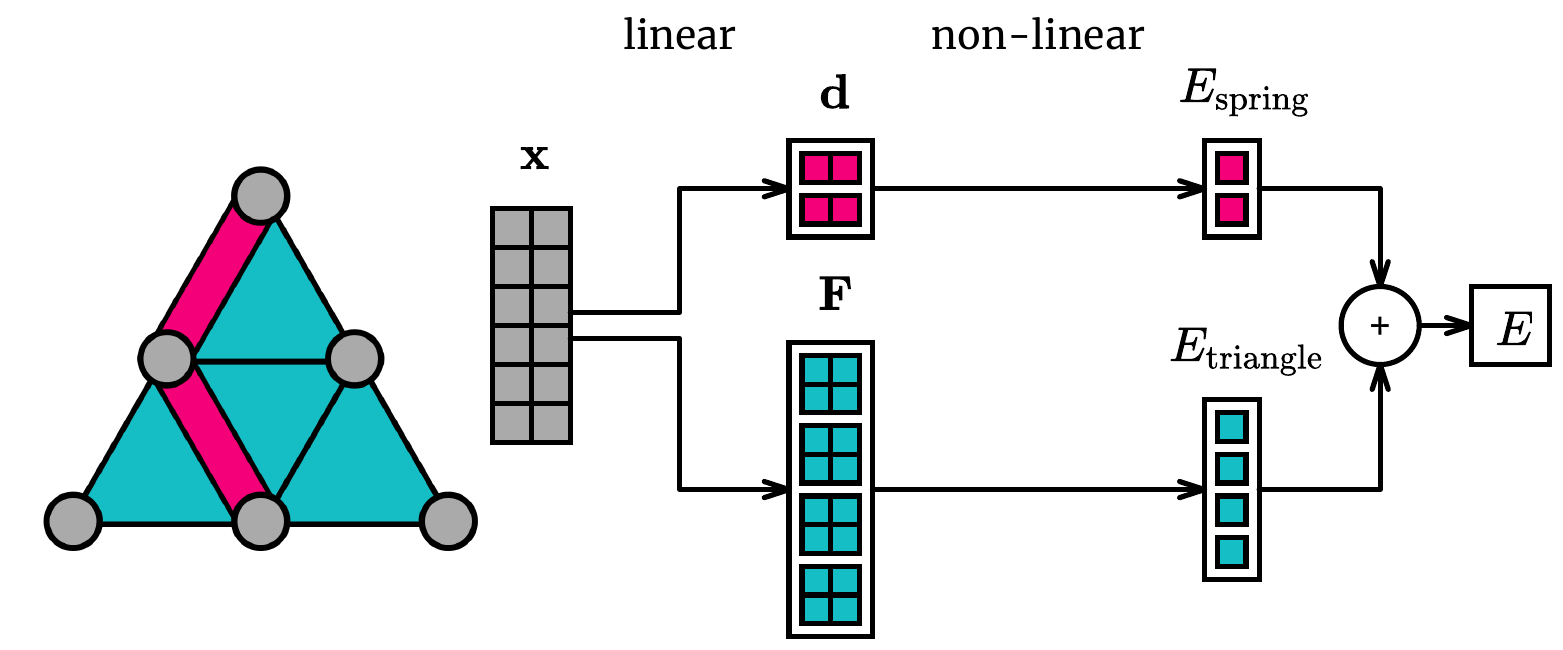}
    \caption{Energy function as a computation graph. Many energy models used in FEM simulation are defined as a non-linear function of the deformation gradient ($\vc{F_i}$ is the deformation gradient of triangle $i$), which is a linear function of the vertex positions $\vc{x}$. Similarly, many energy models used for springs are defined as a non-linear function of $\vc{d}$ (${\vc{d_i}}$ is the vector between the positions of the two vertices connected by spring $i$), which is a linear function of $\vc{x}$.}
    \label{fig:energy-comp-graph}
\end{figure}

Although the specifics may have some differences, there is a high-level interpretation of the energy functions commonly used in soft-body physics that can reveal some of their similarities with neural networks, which are not always evident in more conventional presentations of this topic in the context of continuum mechanics and partial differential equations. Many energy models used in FEM simulation such as Neo-Hookean, Saint Venant-Kirchhoff and corotated elasticity are defined as a non-linear function of the deformation gradient $\vc{F}$, a square matrix (2x2 for triangles and 3x3 for tetrahedra) that can be defined as a linear function of the vertex positions $\vc{x}$. In that sense, all these models can be understood as shallow neural networks that use different non-linearities. We can also see a similar pattern in spring energy functions. \refig{energy-comp-graph} shows a diagram that makes this interpretation more evident.

One difference with more conventional neural network architectures is that the non-linearity in FEM models is not element-wise. For example, an FEM model on 2D triangle meshes computes the energy of every triangle by applying a non-linear function on 2x2 matrices. This approach has some similarities with capsule networks \cite{capsule-nets}, since the non-linearity acts on groups of neurons whose values represent properties of the same entity (one triangle of the mesh), and a similar observation can be made about spring energy functions.

In fact, the energy function used in our simulator is implemented using differentiable operations commonly available in deep learning packages (we used PyTorch \cite{pytorch2019}, without any additional libraries). This also allows us to automatically differentiate the energy function which is useful to implement the forward pass that requires running a minimization procedure (\refeq{x1-quasistatic-argmin}).

Note, however, that the fact that we can implement the energy function using differentiable operations does not immediately provide us with a backward pass for the simulator. Although explicitly unrolling a minimization procedure (for example, gradient descent with a fixed step size) could in principle allow us to run the backward pass, this is not necessarily the best option, since it requires storing in memory all the intermediate steps of the optimizer and limits its applicability to optimization methods that only use differentiable operations in their implementation.

In the next section, we present an implicit differentation approach that allows us to differentiate through the simulator regardless of how the optimization procedure is implemented, it does not require storing intermediate steps and is applicable even if the optimizer runs non-differentiable operations such as line search, which is common in implementations of soft-body physics \cite{descent-methods-gpu, gast-optim, liu2017quasi, stable-neohookean}.

%% file: 03-differentiable-quasistatic.tex
\section{Differentiable quasistatic motions}
\label{sec:diff-qs}

\begin{figure}[h]
    \centering
    \includegraphics[width=\linewidth]{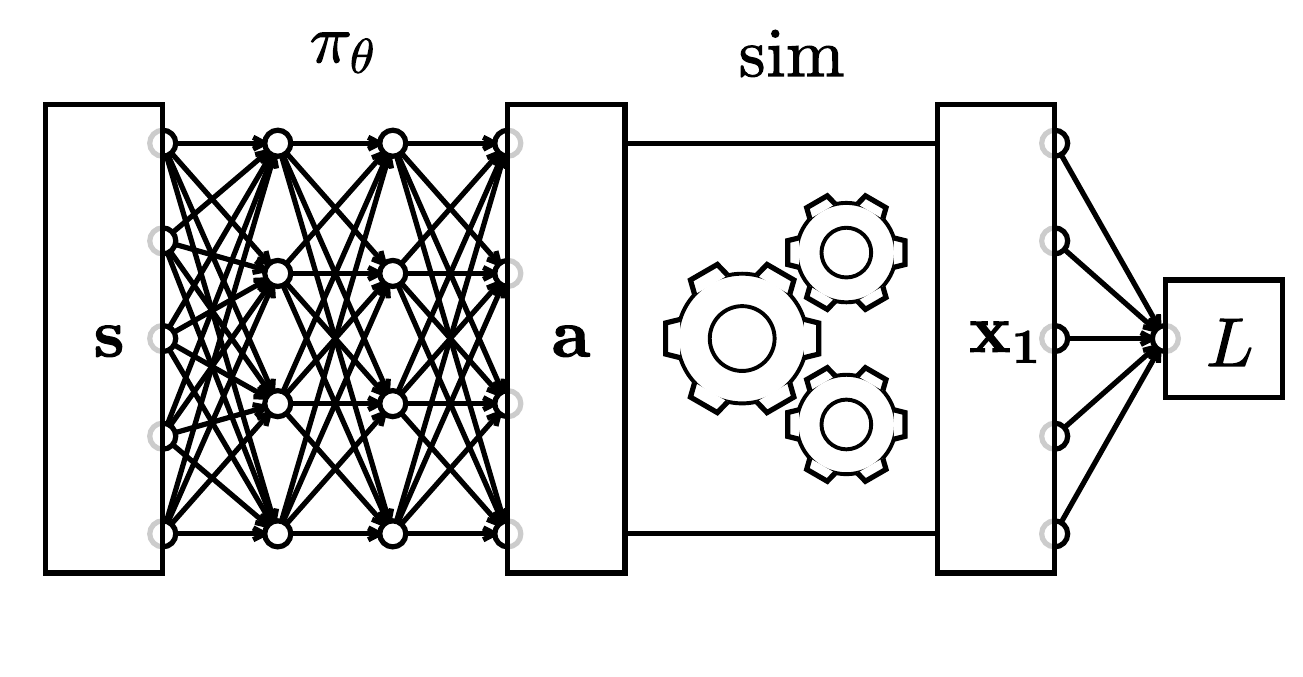}
    \caption{For machine learning problems, we are interested in incorporating physics into neural networks as a layer. For example, we can consider problems where the actions $\vc{a}$ are produced by a neural network policy $\pi_\theta$, and the output of the quasistatic simulator $\vc{x_1}$ is fed into a loss that we want to minimize.}
    \label{fig:comp-to-loss}
\end{figure}

For machine learning problems, we are interested in incorporating physics into neural networks as a layer, as shown in \refig{comp-to-loss}.
The difficulty in running backpropagation on the computation graph presented in \refig{comp-to-loss} is that $\frac{d L}{d \vc a}$ is not readily available because $\vc{x_1}$ is not defined explicitly as a differentiable function of $\vc a$.
In this section, we present an implicit differentiation approach to compute this derivative, and we point out some aspects that make this approach feasible to implement in reverse-mode automatic differentiation systems commonly used for neural network training.
To present an initial derivation using conventional matrix operations, we assume vector representations using one-dimensional arrays for the vertex positions $\vc{x} \in \R^{2n}$ (where $n$ is the number of vertices) and the action $\vc{a} \in \R^{a}$ (where $a$ is the number of actuators), but we show how to generalize this derivation to support other representations for $\vc{x}$ and $\vc{a}$ in \refsec{matrix-free}.

The output of the simulation layer is defined implicitly via energy minimization, which implies that the force $\vc{f} \in \R^{2n}$ (\refeq{f-as-grad}) at the solution is zero (\refeq{f-constraint}).

\begin{equation}
\vc{f}(\vc x, \vc a) = -\nabla_{\vc x} E(\vc x, \vc a)
\label{eq:f-as-grad}
\end{equation}

\begin{equation}
\vc{f}(\vc{x_1}(\vc a), \vc a) = 0
\label{eq:f-constraint}
\end{equation}

We observe that the fact that $\vc{f}$ is zero for any $\vc{a}$ (\refeq{f-constraint}) implies that $\frac{d \vc{f}}{d {\vc a}}$ is also zero for any $\vc{a}$. Applying the chain rule, we get:

\begin{equation}
    \underbrace{
        \dd{}{\vc a}\vc{f}(\vc{x_1}(\vc a), \vc a)
    }_{\R^{2n \times a}} =
    \underbrace{
        \underbrace{
            \left.\frac{\partial \vc{f}}{\partial {\vc x}}\right|_{(\vc{x_1}(\vc a),\vc a)}
        }_{\R^{2n \times 2n}}
        \underbrace{
            \frac{d \vc{x_1}}{d {\vc a}}
        }_{\R^{2n \times a}}
    }_{\R^{2n \times a}} +
    \underbrace{
        \left.\frac{\partial \vc{f}}{\partial {\vc a}}\right|_{(\vc{x_1}(\vc a),\vc a)}
    }_{\R^{2n \times a}} = 0
\label{eq:dfda-v2}
\end{equation}

In the next equations, we will omit the explicit evaluations at $(\vc{x_1}(\vc a), \vc a)$ for notational convenience. From \refeq{dfda-v2} we can obtain a formula for $\dd{\vc{x_1}}{\vc a}$:

\begin{equation}
\dd{\vc{x_1}}{\vc a} =
    - \dpartial{\vc{f}}{\vc x}^{-1}
    \dpartial{\vc{f}}{\vc a}
\label{eq:dx1da}
\end{equation}

Considering the loss $L$ as a function of $\vc x$, we can use \refeq{dx1da} to obtain a formula for $\dd{L}{\vc a}$:

\begin{equation}
\dd{L}{\vc a} = \dd{L}{\vc x} \dd{\vc{x_1}}{\vc a} =
    - \dd{L}{\vc x}
    \dpartial{\vc{f}}{\vc x}^{-1}
    \dpartial{\vc{f}}{\vc a}
\label{eq:dLda}
\end{equation}

We then reorder the matrix multiplications of \refeq{dLda} to get our final formula for the backward pass:

\begin{equation}
\underbrace{\dd{L}{\vc a}}_{\R^{1 \times a}} =
- \underbrace{
\left(
    \underbrace{\dpartial{\vc{f}}{\vc x}^{-1}}_{\R^{2n \times 2n}}
    \underbrace{\dd{L}{\vc x}^{\T}}_{\R^{2n \times 1}}
\right)^{\T}
}_{\R^{1 \times 2n}}
\underbrace{
    \dpartial{\vc{f}}{\vc a}
}_{\R^{2n \times a}}
\label{eq:dLda-re-ordered}
\end{equation}

The derivation of the last equation leveraged the fact that $-\frac{\partial \vc{f}}{\partial\vc x}$ is symmetric, since it is equal to the Hessian of the energy as can be seen via \refeq{f-as-grad}. This Hessian is positive definite when evaluated at a quasistatic configuration $\vc{x_1}(\vc a)$, since it corresponds to a local minimum of the energy $E$.
Thus, the quantity inside the parenthesis in \refeq{dLda-re-ordered}, which requires solving a linear system, can be conveniently computed via a conjugate gradient solver in a matrix-free fashion. The fact that we can implement \refeq{dLda-re-ordered} without explicitly constructing the Hessian matrix is particularly important in the context of neural networks. Neural networks are often implemented using reverse-mode automatic differentiation systems which do not provide an efficient way to compute Hessian matrices, but can compute Hessian-vector products in a matrix-free fashion, which is sufficient to implement \refeq{dLda-re-ordered} (we used PyTorch). We provide more details of this matrix-free implementation in \refsec{matrix-free}.

\subsection{Matrix-free second-order differentiation via backpropagation}
\label{sec:matrix-free}

The formula for the backward pass (\refeq{dLda-re-ordered}) includes $\frac{\partial \vc{f}}{\partial \vc{x}}$ and $\frac{\partial \vc{f}}{\partial \vc{a}}$. We can interpret these quantities as force Jacobian matrices, or matrices that contain second derivatives of the energy function $E$. However, the matrix interpretation of these quantities is not always the most convenient for implementation purposes. In fact, to implement \refeq{dLda-re-ordered} we do not need to construct these matrices, we only need a way to evaluate matrix-vector products.

For example, the quantity inside the parenthesis in \refeq{dLda-re-ordered} requires solving a linear system that can be implemented using a conjugate gradient solver if we can evaluate matrix-vector products $\frac{\partial \vc{f}}{\partial \vc{x}} \vc{v}$ for arbitrary vectors $\vc{v}$. This can be done using the following identity:

\begin{equation}
    \underbrace{
        \underbrace{
            \frac{\partial \vc{f}}{\partial \vc{x}}
        }_{\R^{2n \times 2n}}
        \underbrace{
            \vc{v}
        }_{\R^{2n}}
    }_{\R^{2n}}
    =
    \underbrace{
        \frac{\partial}{\partial \vc{x}}
        \underbrace{
            \langle
                \underbrace{\vc{f}}_{\R^{2n}},
                \underbrace{\vc{v}}_{\R^{2n}}
            \rangle
        }_{\R}
    }_{\R^{2n}}
\label{eq:matrix-free-dfdx}
\end{equation}

This requires creating the force $\vc{f}$ as a differentiable node in the computation graph (for example, in PyTorch, this can be done by running \texttt{torch.autograd.grad} on the energy $E$), and then running backpropagation on the dot product $\langle \vc{f}, \vc{v} \rangle$. Since the dot product is a scalar, running reverse-mode automatic differentiation (backpropagation) on it can be performed efficiently, and it effectively computes the matrix-vector product $\frac{\partial \vc{f}}{\partial \vc{x}} \vc{v}$ that we need. A similar approach can be used for $\frac{\partial \vc{f}}{\partial \vc{a}}$ in order to compute the full expression presented in \refeq{dLda-re-ordered}.

We originally assumed vectors represented as one-dimensional arrays for the action $\vc{a} \in \R^{a}$ and vertex positions $\vc{x} \in \R^{2n}$ to present the initial derivation using conventional matrix operations, but with a matrix-free approach it becomes more clear that we do not have to restrict these quantities to one-dimensional arrays. A matrix is just one particular representation of a linear function and matrix-vector multiplication is just one particular implementation of linear function application. Instead of interpreting $\frac{\partial \vc{f}}{\partial \vc{x}}$ as a matrix, we can interpret it as a linear function that takes as input one argument of the same type of $\vc{x}$ and outputs a value of the same type of $\vc{x}$ ($\R^{2n}$ is one possible choice, but not the only one).

We can implement this for arbitrary representations of $\vc{x}$ (as well as $\vc{a}$) if we can define the appropriate dot product used in \refeq{matrix-free-dfdx} to compute second derivatives in a matrix-free fashion. The dot product can be easily defined in general for multidimensional arrays of arbitrary order (or tensors, as commonly referred to in deep learning packages) via an element-wise multiplication followed by a sum of all the elements.
In fact, as suggested by Figures \ref{fig:qs-action} and \ref{fig:energy-comp-graph}, vertex positions $\vc{x}$ are more naturally represented as an array of two dimensions, where one dimension indexes vertices and the other indexes spatial dimensions.
This and other multidimensional array representations are naturally handled by our matrix-free approach without any modifications since we do not require $\vc{x}$ to be a one-dimensional array nor do we require constructing $\frac{\partial \vc{f}}{\partial \vc{x}}$ explicitly as a matrix.

%% file: 04-dynamics.tex
\section{Differentiable implicit numerical integration}
\label{sec:dynamics}

We now consider the effect of inertia, which we previously ignored in quasistatic motions. For this purpose, we consider sequences of states $\vc{s_t}$ that evolve over time. The state of our simulation is defined by vertex positions and velocities $\vc{s_t} = (\vc{x_t}, \vc{v_t})$. A good starting point to characterize the dynamics of our system is to consider the equations of motion in a continuous-time setting:

\begin{equation}
    \label{eq:true-integration}
    \begin{split}
    \vc{x}(t + h) &= \vc{x}(t) + \int_{t}^{t + h} \vc{v}(t) dt \\
    \vc{v}(t + h) &= \vc{v}(t) + \vc{M}^{-1} \int_{t}^{t + h} \vc{f}(\vc{x}(t)) dt
    \end{split}
\end{equation}

where $\vc{f}$ is the force and $\vc{M}$ is the mass. For simulation purposes, we assume a particular time discretization scheme given by a numerical integration method, defining a function $\mathrm{sim}$ that updates the state of the system after $h$ units of time:

\begin{equation}
    \vc{s_1} = \mathrm{sim}(\vc{s_0}, \vc{a_0})
    \label{eq:sim-dynamics}
\end{equation}

Explicit methods such as symplectic Euler have been used before for differentiable soft-body physics \cite{hu2019chainqueen, difftaichi}. However, in this paper we focus on implicit numerical integration methods that do not provide an explicit differentiable forward pass. Although implicit numerical integration methods are sometimes formulated as root finding problems using the force function, many implicit numerical integration methods can be reformulated as minimization problems using the energy function. The latter formulation is preferred in practical implementations due to its better numerical properties \cite{martin2011example, gast-optim, liu2017quasi, rojas2018-avf-stvk}. We adopt the optimization-based formulation of implicit numerical integration for our differentiable simulator. The general form that these methods have is the following:

\begin{equation}
    \begin{split}
    \vc{x_1} &= \argmin{\vc x}{g(\vc{x}, \vc{s_0}, \vc{a_0})} \\
    \vc{v_1} &= G(\vc{s_0}, \vc{x_1})
    \end{split}
    \label{eq:opt-integration}
\end{equation}

\begin{figure}[h]
    \centering
    \includegraphics[width=0.73\linewidth]{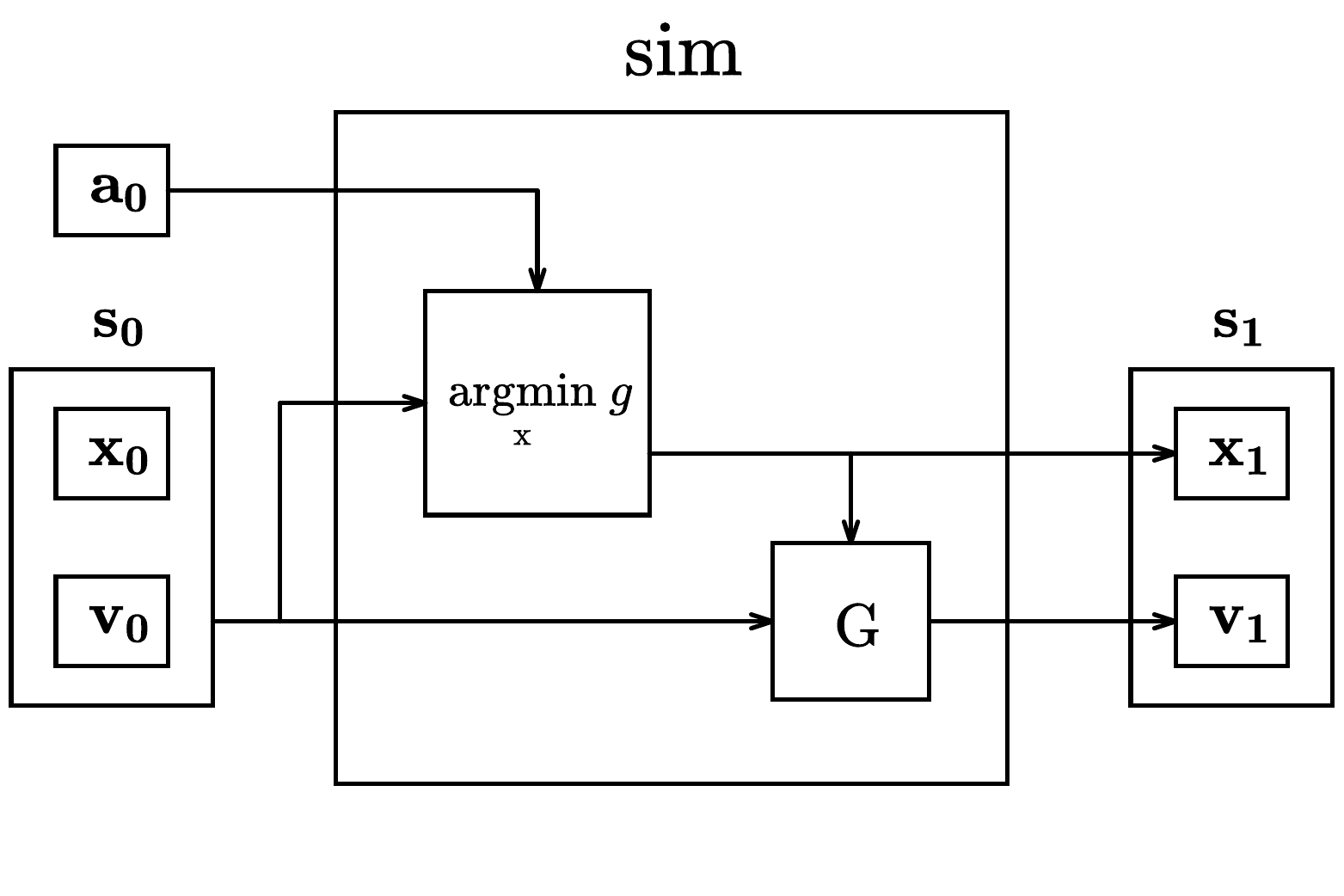}
    \caption{The differentiable implicit numerical integration layers that we consider consist of two steps: an implicit update rule (finding a local minimum of $g$) to compute vertex positions $\vc{x_1}$, followed by an explicit update rule $G$ to compute vertex velocities $\vc{v_1}$.}
    \label{fig:dynamics-cell}
\end{figure}

where $g$ is a scalar-valued function that includes the potential energy $E$, and $G$ is an explicit velocity update rule. For example, the optimization-based formulation of backward Euler integration commonly used in simulation \cite{gast-optim, liu2017quasi} can be written as:

\begin{equation}
    \begin{split}
    \vc{x_1} &= \argmin{\vc x}{\frac{1}{2}\|\vc{x} - (\vc{x_0} + h \vc{v_0})\|_{\vc{M}}^2 + h^2 E({\vc x}, \vc{a_0})} \\
    \vc{v_1} &= \frac{\vc{x_1} - \vc{x_0}}{h}
    \end{split}
    \label{eq:opt-be}
\end{equation}

The implicit update rule for $\vc{x_1}$ can be differentiated using the same implicit differentiation strategy presented in \refsec{diff-qs} and the explicit update rule for $\vc{v_1}$ can be differentiated using standard backpropagation. This provides us with a fully-differentiable implicit numerical integration layer (\refig{dynamics-cell}) that we can use for policy optimization.

\subsection{Locomotion}

To demonstrate the effectiveness of our proposed implicit numerical integration layer, we consider episodic tasks where an agent gets a reward at the end of the episode equal to its total horizontal displacement. We used backward Euler integration for our simulation and introduced energy terms to model collisions and friction with a flat terrain (see \refsec{collision-and-friction} for details). Since every step of the simulation is differentiable, we can run backpropagation through time to optimize control policies. Our policies are modeled as neural networks with a single hidden layer with 32 ReLU units that take as input the current state of the simulation (vertex positions and velocities) and output actions that change the rest length of the contractile fibers.

\begin{figure}[h]
    \centering
    \includegraphics[width=\linewidth]{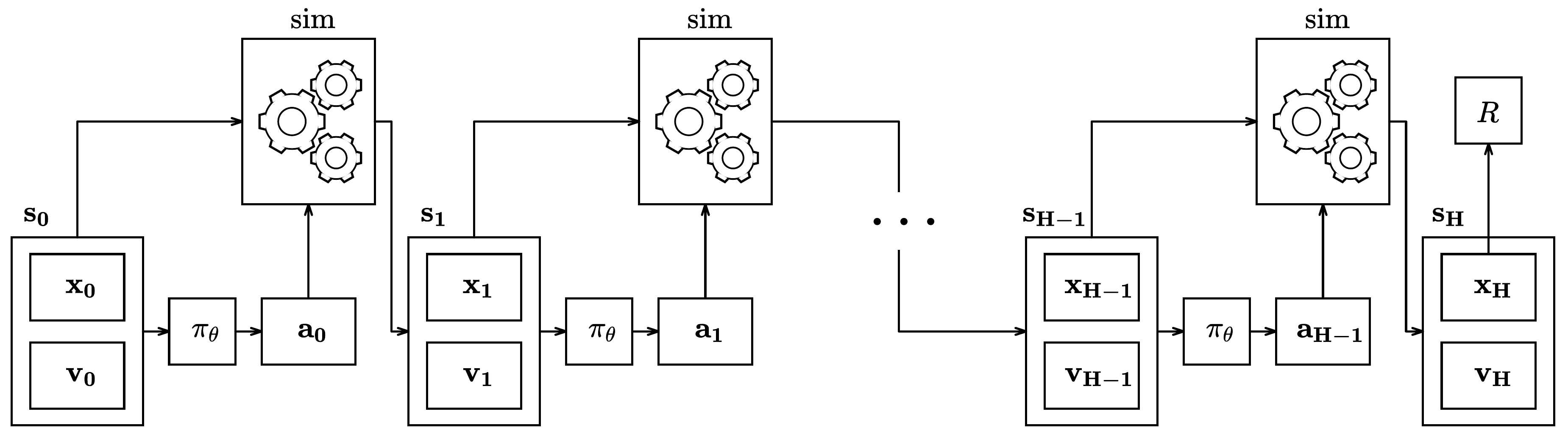}
    \caption{We consider episodic tasks with a fixed time horizon $H$ where the agent gets a reward $R$ at the end of the episode equal to its total horizontal displacement. Since every step of the simulation is differentiable, we can run backpropagation through time to optimize control policies $\pi_\theta$.}
    \label{fig:dynamics-seq}
\end{figure}

\begin{figure}[h]
    \centering
    \includegraphics[width=\linewidth]{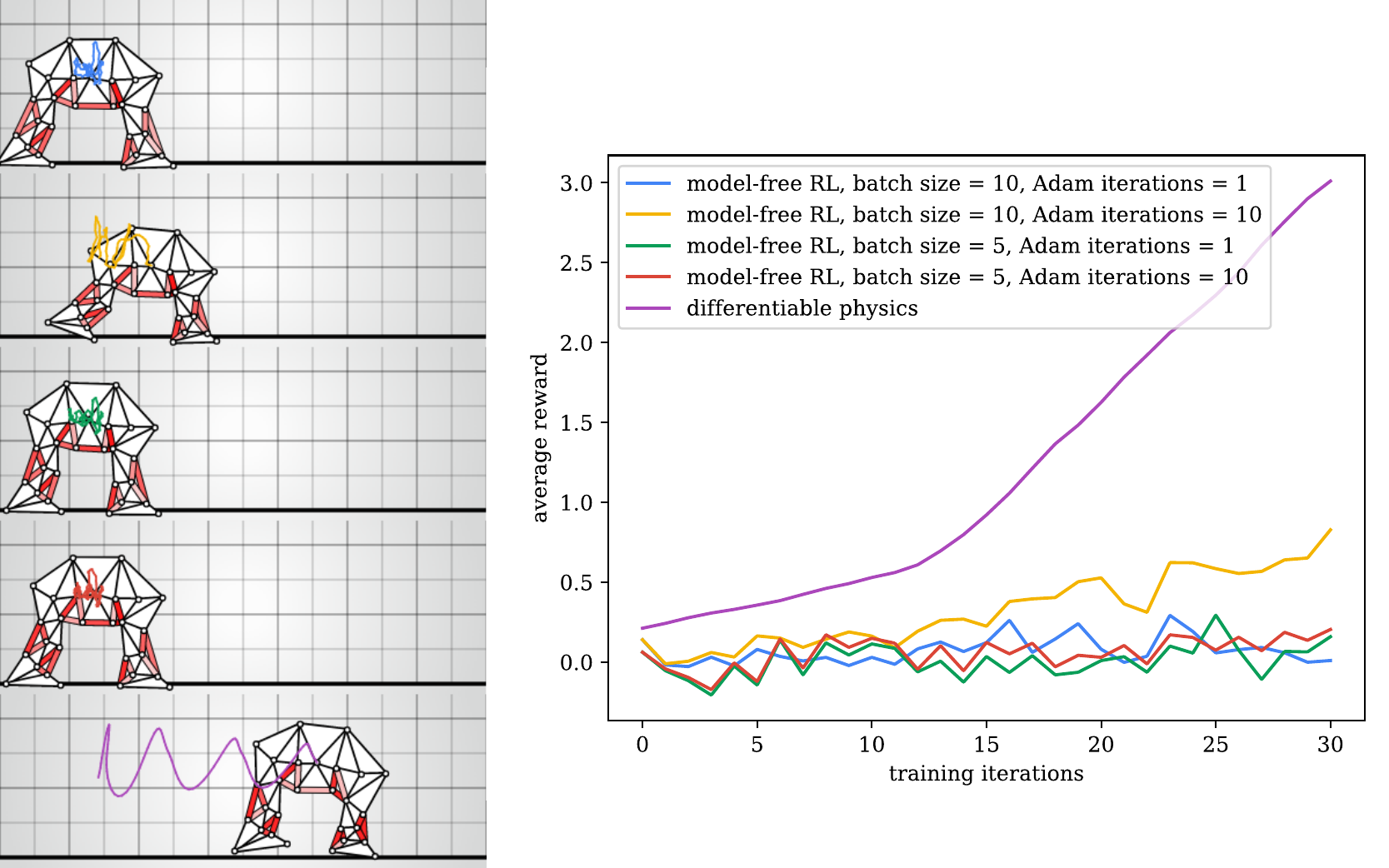}
    \caption{Our differentiable physics approach achieves better sample efficiency than a model-free RL approach based on PPO across multiple hyperparameter choices. The left shows the result of evaluating the policy for 100 time steps, after running 30 iterations of different training procedures. The plot (right) shows the average reward achieved by the agent in a fixed time horizon of 100 time steps, after every training iteration. Note that one training iteration of model-free RL performs multiple Adam iterations using multiple sampled trajectories (batch size), while our differentiable simulator performs a single Adam iteration with an effective batch size of 1 per training iteration.}
    \label{fig:hdog1-single-plot}
\end{figure}

\begin{figure}[h!]
    \centering
    \includegraphics[width=\linewidth]{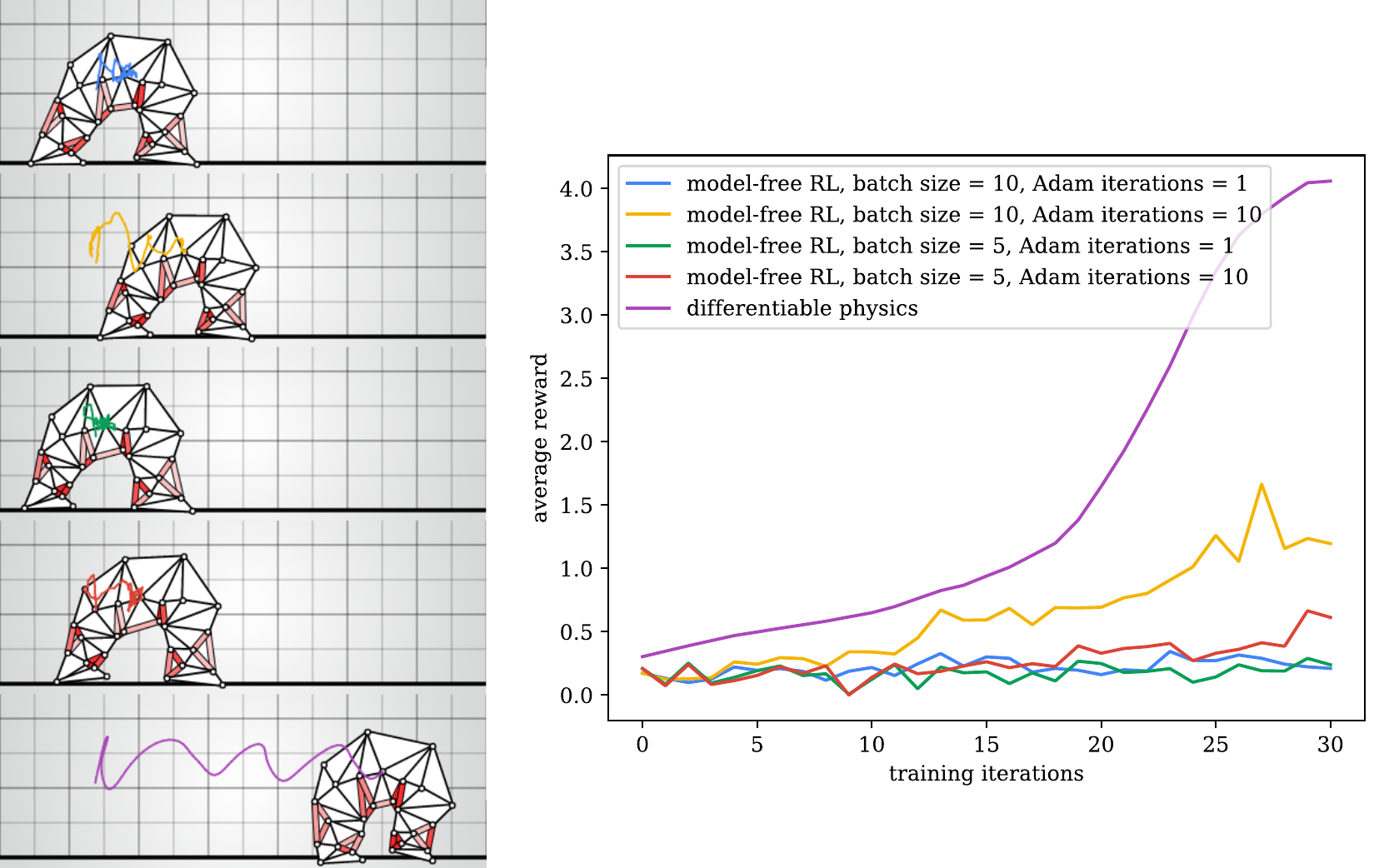}
    \caption{Additional runs of the experiment shown in \refig{hdog1-single-plot}, using a different agent.}
    \label{fig:hdog2-single-plot}
\end{figure}

We compared the performance of our method with model-free reinforcement learning. We used the same training strategy based on PPO \cite{ppo} used in \cite{rojas2019-drl-sbl} for soft-body locomotion (clipping parameter $\epsilon = 0.2$) and found that, overall, our differentiable simulator requires fewer iterations and samples to make progress, as shown in \refig{hdog1-single-plot} and \refig{hdog2-single-plot}. Note that PPO requires sampling multiple trajectories to approximate the gradient of the simulator, whose accuracy depends on the number of trajectories sampled every training iteration (batch size) and requires running a stochastic policy, while with a differentiable simulator we can train a deterministic policy using an effective batch size of 1. Also note that one training iteration of PPO actually performs multiple optimization steps (we used Adam \cite{adam} with a learning rate of $10^{-3}$), while our differentiable simulator performs a single Adam iteration per training iteration.

\begin{figure}[h!]
    \centering
    \includegraphics[width=\linewidth]{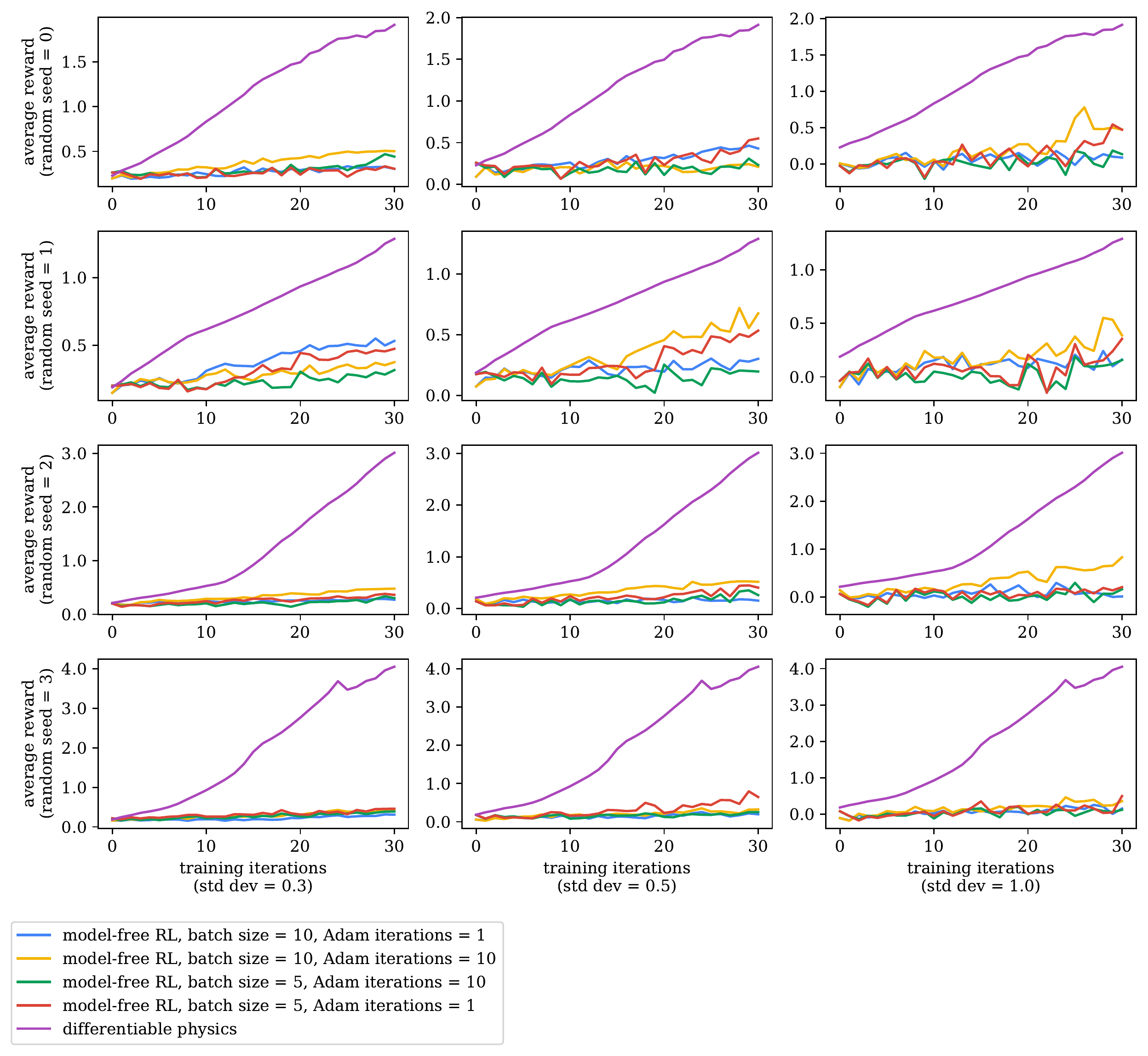}
    \caption{Additional runs of the experiment shown in \refig{hdog1-single-plot} using different random seeds and initial values for the standard deviation used for the stochastic policy. Note that the plots for the differentiable simulator across the same row are equal because the policy is deterministic.}
    \label{fig:hdog1-matrix-plot}
\end{figure}

\begin{figure}[h!]
    \centering
    \includegraphics[width=\linewidth]{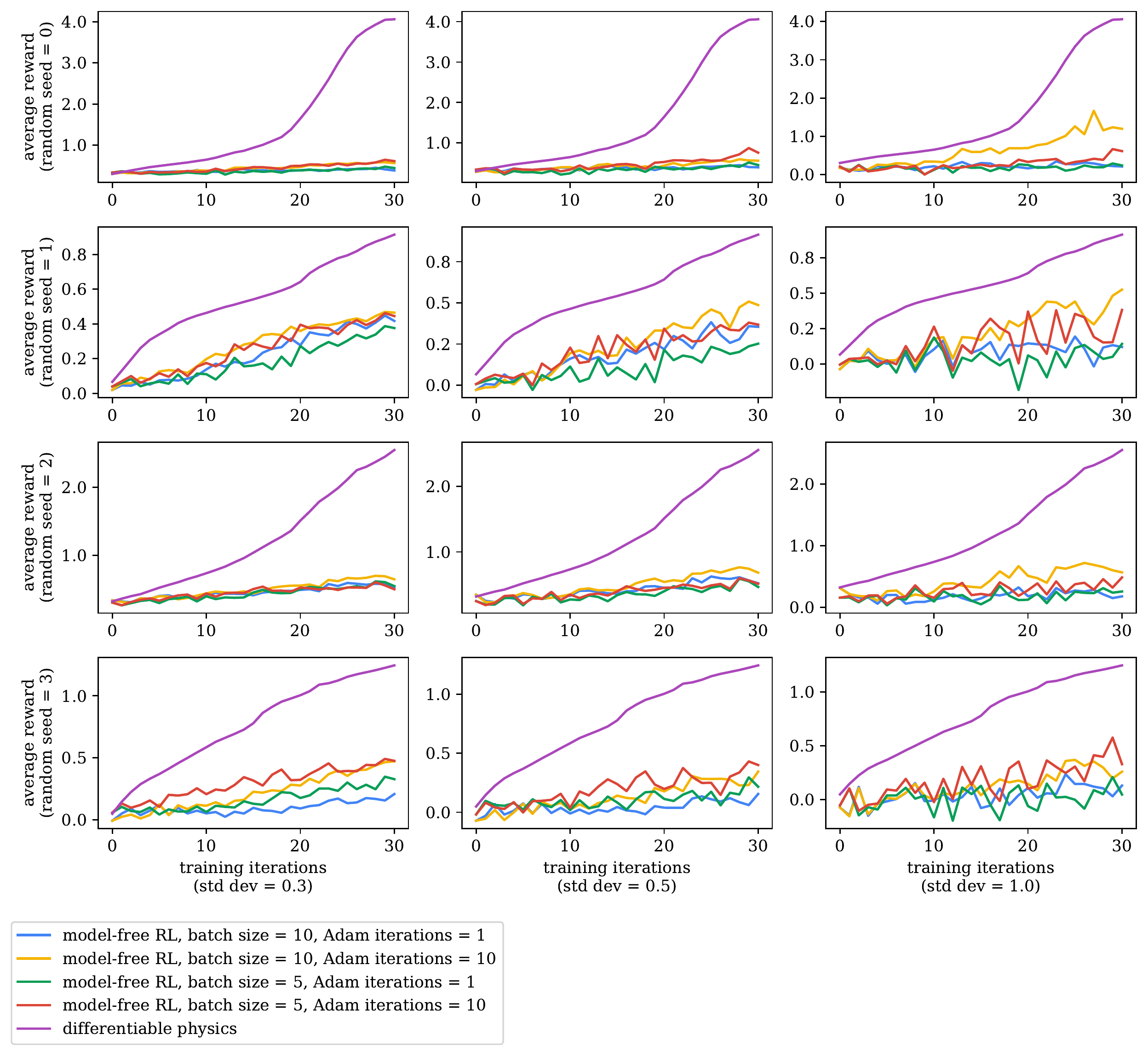}
    \caption{Additional runs of the experiment shown in \refig{hdog2-single-plot} using different random seeds and initial values for the standard deviation used in the stochastic policy. Note that the plots for the differentiable simulator across the same row are equal because the policy is deterministic.}
    \label{fig:hdog2-matrix-plot}
\end{figure}

For model-free RL, we modeled the stochastic policy as a multivariate Gaussian distribution, where the mean is computed using the same architecture of the deterministic policy used for our differentiable simulator, and they were both initialized with the same parameters. The stochastic policy samples actions from this distribution, using a trainable standard deviation for every muscle fiber, which is initialized with the same value for all fibers.
We also repeated these experiments with multiple random seeds and different initialization values for the standard deviation of the stochastic policy (\refig{hdog1-matrix-plot} and \refig{hdog2-matrix-plot}) and found that the differentiable simulator can also make progress in fewer training iterations across multiple runs.

\subsection{Ground collisions and friction}
\label{sec:collision-and-friction}

To handle collisions with the ground, we implemented a one-sided quadratic energy term. Assuming a flat terrain that coincides with the horizontal axis, as shown in our experiments, we can define a collision energy using the ReLU function to introduce a penalty term (scaled by a stiffness coefficient $k_{\mathrm{collision}}$) that is only active when a vertex with position $\vc{x} = (\vc{x}_x, \vc{x}_y)$ is in contact with the ground:

\begin{equation}
E_{\mathrm{collision}}(\vc{x}) = k_{\mathrm{collision}} (\mathrm{ReLU}(-\vc{x}_y))^2
\label{eq:energy-collision}
\end{equation}

We also implemented friction between the deformable mesh and the ground to enable locomotion. Conceptually, the value of the friction term should depend on the horizontal component of the velocity. Hence, we introduce a velocity function $\vc{v}$ that can be expressed in terms of vertex positions and is consistent with the explicit velocity update rule used in backward Euler:

\begin{equation}
\vc{v}(\vc{x}) = \frac{\vc{x} - \vc{x_0}}{h}
\label{eq:v-of-x-vec}
\end{equation}

The final value of the friction term we used also depends on the amount of overlap between the ground and the vertical component of the vertex position in the previous time step $\vc{x_0}_y$ (using a numerical tolerance $\epsilon$):

\begin{equation}
E_{\mathrm{friction}}(\vc{x}) = k_{\mathrm{friction}} (\vc{v}_x (\vc{x}))^2 \mathrm{ReLU}(-\vc{x_0}_y + \epsilon)
\label{eq:energy-friction}
\end{equation}

The effect that this energy term has is that it penalizes high speeds in the horizontal plane ($\vc{v}_x (\vc{x})$) for vertices that are in contact with the ground (determined by $\vc{x_0}_y$). Since this term is part of the function that the simulator minimizes in the forward pass, it naturally tends to reduce the speed of vertices that are in contact with the ground over time, effectively modeling a type of friction. As with the other energy functions used in our simulation, this model can be easily changed to accomodate different friction models, since our energy-based formulation detailed in \refsec{diff-qs} allows us to compute all the required derivatives automatically via backpropagation provided that the energy function is implemented using differentiable operations.

%% file: 05-conclusion.tex
\section{Conclusion and future work}

We presented a soft-body physics layer that can be differentiated even when an explicit differentiable forward pass is not available. By adopting an optimization-based approach to define state transitions, we derived a differentiation rule that is agnostic to the specific implementation of the forward pass, which allows us to achieve differentiability for implicit numerical integration methods.
We showed how our implicit differentiation approach can be implemented using reverse-mode automatic differentiation, which allows us to implement and compose our simulator with neural networks using the same automatic differentiation system.

As with recent work in differentiable physics for explicit methods, such as DiffTaichi \cite{difftaichi}, one interesting direction of future work for implicit methods is to devise performance optimizations for the simulator while keeping the same flexibility and productivity provided by automatic differentiation. Our current implementation does not take full advantage of spatial sparsity, which is one of the main reasons why simulators implemented in DiffTaichi are often faster than implementations in conventional deep learning packages. The forward pass of our simulator can also be further improved by adopting different numerical optimization strategies. We adopted a descent method with line search that only uses information of first-order derivatives, but this could be improved in the future. Our goal in this paper is not to provide a specific implementation of the forward pass, but rather to present a general approach that leverages reverse-mode automatic differentiation and allows us to use different implementations of the forward pass while keeping the simulator differentiable.

We demonstrated that our differentiable simulator is compatible with backpropagation through time and can be used to optimize control policies taking into account the effect of multiple simulation steps. However, as with other recurrent models, vanishing and exploding gradients could become an issue for very long time horizons.
We ran some preliminary experiments that indicate that combining our differentiable simulator with a trainable value function can mitigate this issue, which is also common in infinite-horizon reinforcement learning, but further research is needed to properly assess the benefits of this approach.
Although a differentiable physics engine allows us to optimize deterministic policies, it is possible that adding some stochasticity, for example to improve exploration during training, could mitigate potential issues with local optima, which has been reported in prior work in differentiable physics.
We believe it is worth evaluating the benefits of incorporating these modifications into the training loop in future work.

%% file: acknowledgements.tex
\section*{Acknowledgements}
We would like to thank Alexander Winkler for the useful discussions.
This material is based upon work supported by the National Science Foundation under Grant Numbers IIS-2008915, IIS-2008584, IIS-1763638 and IIS-1764071. Any opinions, findings, and conclusions or recommendations expressed in this material are those of the author(s) and do not necessarily reflect the views of the National Science Foundation.

%% file: main.bbl
\begin{thebibliography}{27}
\providecommand{\natexlab}[1]{#1}
\providecommand{\url}[1]{\texttt{#1}}
\expandafter\ifx\csname urlstyle\endcsname\relax
  \providecommand{\doi}[1]{doi: #1}\else
  \providecommand{\doi}{doi: \begingroup \urlstyle{rm}\Url}\fi

\bibitem[Agrawal et~al.(2019)Agrawal, Amos, Barratt, Boyd, Diamond, and
  Kolter]{cvxpylayers2019}
Agrawal, A., Amos, B., Barratt, S., Boyd, S., Diamond, S., and Kolter, Z.
\newblock Differentiable convex optimization layers.
\newblock In \emph{Advances in Neural Information Processing Systems}, 2019.

\bibitem[Amos \& Kolter(2017)Amos and Kolter]{amos2017optnet}
Amos, B. and Kolter, J.~Z.
\newblock {O}pt{N}et: Differentiable optimization as a layer in neural
  networks.
\newblock In \emph{Proceedings of the 34th International Conference on Machine
  Learning}, volume~70 of \emph{Proceedings of Machine Learning Research}, pp.\
   136--145. PMLR, 2017.

\bibitem[Bai et~al.(2019)Bai, Kolter, and Koltun]{deqs}
Bai, S., Kolter, J.~Z., and Koltun, V.
\newblock Deep equilibrium models.
\newblock In Wallach, H., Larochelle, H., Beygelzimer, A., d\textquotesingle
  Alch\'{e}-Buc, F., Fox, E., and Garnett, R. (eds.), \emph{Advances in Neural
  Information Processing Systems 32}, pp.\  690--701. Curran Associates, Inc.,
  2019.
\newblock URL
  \url{http://papers.nips.cc/paper/8358-deep-equilibrium-models.pdf}.

\bibitem[Bern et~al.(2017{\natexlab{a}})Bern, Kumagai, and
  Coros]{bern-fabrication-modeling-control-plush-robots}
Bern, J., Kumagai, G., and Coros, S.
\newblock Fabrication, modeling, and control of plush robots.
\newblock In \emph{2017 IEEE/RSJ International Conference on Intelligent Robots
  and Systems (IROS)}, pp.\  3739 -- 3746, Piscataway, NJ, 2017{\natexlab{a}}.
  IEEE.
\newblock ISBN 978-1-5386-2682-5.
\newblock \doi{10.1109/IROS.2017.8206223}.
\newblock 2017 IEEE/RSJ International Conference on Intelligent Robots and
  Systems (IROS 2017); Conference Location: Vancouver, Canada; Conference Date:
  September 24-28, 2017.

\bibitem[Bern et~al.(2017{\natexlab{b}})Bern, Chang, and
  Coros]{animated-plushies}
Bern, J.~M., Chang, K.-H., and Coros, S.
\newblock Interactive design of animated plushies.
\newblock \emph{ACM Trans. Graph.}, 36\penalty0 (4):\penalty0 80:1--80:11, July
  2017{\natexlab{b}}.
\newblock ISSN 0730-0301.
\newblock \doi{10.1145/3072959.3073700}.
\newblock URL \url{http://doi.acm.org/10.1145/3072959.3073700}.

\bibitem[Bern et~al.(2019)Bern, Banzet, Poranne, and Coros]{bern2019trajectory}
Bern, J.~M., Banzet, P., Poranne, R., and Coros, S.
\newblock Trajectory optimization for cable-driven soft robot locomotion.
\newblock \emph{Proceedings of Robotics: Science and Systems,
  FreiburgimBreisgau, Germany}, 2019.

\bibitem[Chen et~al.(2018)Chen, Rubanova, Bettencourt, and
  Duvenaud]{neural-odes}
Chen, R. T.~Q., Rubanova, Y., Bettencourt, J., and Duvenaud, D.~K.
\newblock Neural ordinary differential equations.
\newblock In Bengio, S., Wallach, H., Larochelle, H., Grauman, K.,
  Cesa-Bianchi, N., and Garnett, R. (eds.), \emph{Advances in Neural
  Information Processing Systems}, volume~31, pp.\  6571--6583. Curran
  Associates, Inc., 2018.
\newblock URL
  \url{https://proceedings.neurips.cc/paper/2018/file/69386f6bb1dfed68692a24c8686939b9-Paper.pdf}.

\bibitem[Degrave et~al.(2019)Degrave, Hermans, Dambre, and
  wyffels]{degrave-differentiable}
Degrave, J., Hermans, M., Dambre, J., and wyffels, F.
\newblock A differentiable physics engine for deep learning in robotics.
\newblock \emph{Frontiers in Neurorobotics}, 13:\penalty0 6, 2019.
\newblock ISSN 1662-5218.
\newblock \doi{10.3389/fnbot.2019.00006}.
\newblock URL
  \url{https://www.frontiersin.org/article/10.3389/fnbot.2019.00006}.

\bibitem[Gast \& Schroeder(2015)Gast and Schroeder]{gast-optim}
Gast, T.~F. and Schroeder, C.
\newblock Optimization integrator for large time steps.
\newblock In \emph{Proceedings of the ACM SIGGRAPH/Eurographics Symposium on
  Computer Animation}, SCA '14, pp.\  31–40, Goslar, DEU, 2015. Eurographics
  Association.

\bibitem[Geilinger et~al.(2020)Geilinger, Hahn, Zehnder, B\"{a}cher,
  Thomaszewski, and Coros]{add-siggraph}
Geilinger, M., Hahn, D., Zehnder, J., B\"{a}cher, M., Thomaszewski, B., and
  Coros, S.
\newblock {ADD}: Analytically differentiable dynamics for multi-body systems
  with frictional contact.
\newblock \emph{ACM Trans. Graph.}, 39\penalty0 (6), November 2020.
\newblock ISSN 0730-0301.
\newblock \doi{10.1145/3414685.3417766}.
\newblock URL \url{https://doi.org/10.1145/3414685.3417766}.

\bibitem[Hu et~al.(2019{\natexlab{a}})Hu, Anderson, Li, Sun, Carr,
  Ragan-Kelley, and Durand]{difftaichi}
Hu, Y., Anderson, L., Li, T.-M., Sun, Q., Carr, N., Ragan-Kelley, J., and
  Durand, F.
\newblock {DiffTaichi}: Differentiable programming for physical simulation,
  2019{\natexlab{a}}.

\bibitem[Hu et~al.(2019{\natexlab{b}})Hu, Li, Anderson, Ragan-Kelley, and
  Durand]{hu2019-taichi-lang}
Hu, Y., Li, T.-M., Anderson, L., Ragan-Kelley, J., and Durand, F.
\newblock Taichi: A language for high-performance computation on spatially
  sparse data structures.
\newblock \emph{ACM Trans. Graph.}, 38\penalty0 (6):\penalty0 201:1--201:16,
  November 2019{\natexlab{b}}.
\newblock ISSN 0730-0301.
\newblock \doi{10.1145/3355089.3356506}.
\newblock URL \url{http://doi.acm.org/10.1145/3355089.3356506}.

\bibitem[Hu et~al.(2019{\natexlab{c}})Hu, Liu, Spielberg, Tenenbaum, Freeman,
  Wu, Rus, and Matusik]{hu2019chainqueen}
Hu, Y., Liu, J., Spielberg, A., Tenenbaum, J.~B., Freeman, W.~T., Wu, J., Rus,
  D., and Matusik, W.
\newblock {ChainQueen}: A real-time differentiable physical simulator for soft
  robotics.
\newblock In \emph{2019 International Conference on Robotics and Automation
  (ICRA)}, pp.\  6265--6271. IEEE, 2019{\natexlab{c}}.

\bibitem[Irving et~al.(2004)Irving, Teran, and Fedkiw]{irving-invertible}
Irving, G., Teran, J., and Fedkiw, R.
\newblock Invertible finite elements for robust simulation of large
  deformation.
\newblock In \emph{Proceedings of the 2004 ACM SIGGRAPH/Eurographics Symposium
  on Computer Animation}, SCA '04, pp.\  131–140, Goslar, DEU, 2004.
  Eurographics Association.
\newblock ISBN 3905673142.
\newblock \doi{10.1145/1028523.1028541}.
\newblock URL \url{https://doi.org/10.1145/1028523.1028541}.

\bibitem[Kingma \& Ba(2015)Kingma and Ba]{adam}
Kingma, D.~P. and Ba, J.
\newblock Adam: {A} method for stochastic optimization.
\newblock In \emph{{ICLR}}, 2015.

\bibitem[LeCun et~al.(2006)LeCun, Chopra, Hadsell, Huang, and
  et~al.]{ebm-tutorial}
LeCun, Y., Chopra, S., Hadsell, R., Huang, F.~J., and et~al.
\newblock A tutorial on energy-based learning.
\newblock In \emph{Predicting Structured Data}. MIT Press, 2006.

\bibitem[Liu et~al.(2017)Liu, Bouaziz, and Kavan]{liu2017quasi}
Liu, T., Bouaziz, S., and Kavan, L.
\newblock {Quasi-Newton} methods for real-time simulation of hyperelastic
  materials.
\newblock \emph{ACM Transactions on Graphics (TOG)}, 36\penalty0 (3):\penalty0
  23, 2017.

\bibitem[Martin et~al.(2011)Martin, Thomaszewski, Grinspun, and
  Gross]{martin2011example}
Martin, S., Thomaszewski, B., Grinspun, E., and Gross, M.
\newblock Example-based elastic materials.
\newblock In \emph{ACM Transactions on Graphics (TOG)}, volume~30, pp.\ ~72.
  ACM, 2011.

\bibitem[Paszke et~al.(2019)Paszke, Gross, Massa, Lerer, Bradbury, Chanan,
  Killeen, Lin, Gimelshein, Antiga, Desmaison, Kopf, Yang, DeVito, Raison,
  Tejani, Chilamkurthy, Steiner, Fang, Bai, and Chintala]{pytorch2019}
Paszke, A., Gross, S., Massa, F., Lerer, A., Bradbury, J., Chanan, G., Killeen,
  T., Lin, Z., Gimelshein, N., Antiga, L., Desmaison, A., Kopf, A., Yang, E.,
  DeVito, Z., Raison, M., Tejani, A., Chilamkurthy, S., Steiner, B., Fang, L.,
  Bai, J., and Chintala, S.
\newblock {PyTorch}: An imperative style, high-performance deep learning
  library.
\newblock In \emph{Advances in Neural Information Processing Systems 32}. 2019.

\bibitem[Rojas et~al.(2018)Rojas, Liu, and Kavan]{rojas2018-avf-stvk}
Rojas, J., Liu, T., and Kavan, L.
\newblock Average vector field integration for {St. Venant-Kirchhoff}
  deformable models.
\newblock \emph{IEEE Transactions on Visualization and Computer Graphics},
  2018.

\bibitem[Rojas et~al.(2019)Rojas, Coros, and Kavan]{rojas2019-drl-sbl}
Rojas, J., Coros, S., and Kavan, L.
\newblock Deep reinforcement learning for 2{D} soft body locomotion.
\newblock In \emph{NeurIPS Workshop on Machine Learning for Creativity and
  Design 3.0}, 2019.

\bibitem[Sabour et~al.(2017)Sabour, Frosst, and Hinton]{capsule-nets}
Sabour, S., Frosst, N., and Hinton, G.~E.
\newblock Dynamic routing between capsules.
\newblock In Guyon, I., Luxburg, U.~V., Bengio, S., Wallach, H., Fergus, R.,
  Vishwanathan, S., and Garnett, R. (eds.), \emph{Advances in Neural
  Information Processing Systems}, volume~30, pp.\  3856--3866. Curran
  Associates, Inc., 2017.
\newblock URL
  \url{https://proceedings.neurips.cc/paper/2017/file/2cad8fa47bbef282badbb8de5374b894-Paper.pdf}.

\bibitem[Schulman et~al.(2017)Schulman, Wolski, Dhariwal, Radford, and
  Klimov]{ppo}
Schulman, J., Wolski, F., Dhariwal, P., Radford, A., and Klimov, O.
\newblock Proximal policy optimization algorithms.
\newblock \emph{CoRR}, abs/1707.06347, 2017.

\bibitem[Sifakis \& Barbič(2012)Sifakis and Barbič]{femdefo}
Sifakis, E. and Barbič, J.
\newblock {FEM} simulation of 3{D} deformable solids: A practitioner's guide to
  theory, discretization and model reduction.
\newblock \url{http://www.femdefo.org}, 2012.

\bibitem[Sifakis et~al.(2005)Sifakis, Neverov, and
  Fedkiw]{sifakis2005automatic}
Sifakis, E., Neverov, I., and Fedkiw, R.
\newblock Automatic determination of facial muscle activations from sparse
  motion capture marker data.
\newblock volume~24, pp.\  417--425, 2005.

\bibitem[Smith et~al.(2018)Smith, Goes, and Kim]{stable-neohookean}
Smith, B., Goes, F., and Kim, T.
\newblock Stable neo-hookean flesh simulation.
\newblock \emph{ACM Transactions on Graphics}, 37:\penalty0 1--15, 03 2018.
\newblock \doi{10.1145/3180491}.

\bibitem[Wang \& Yang(2016)Wang and Yang]{descent-methods-gpu}
Wang, H. and Yang, Y.
\newblock Descent methods for elastic body simulation on the gpu.
\newblock \emph{ACM Trans. Graph.}, 35\penalty0 (6), November 2016.
\newblock ISSN 0730-0301.
\newblock \doi{10.1145/2980179.2980236}.
\newblock URL \url{https://doi.org/10.1145/2980179.2980236}.

\end{thebibliography}
